\pdfoutput=1

\documentclass[11pt]{article}

\usepackage[]{acl}

\usepackage{times}
\usepackage{latexsym}
\usepackage{amssymb}
\usepackage{graphicx}
\usepackage{amsmath}
\usepackage{algorithm}
\usepackage{algpseudocode}
\usepackage{booktabs} 
\usepackage{xcolor}
\usepackage{todonotes}
\usepackage{array}
\usepackage{subcaption}
\usepackage[T1]{fontenc}
\usepackage{xspace}
\usepackage{graphicx}
\usepackage{subcaption}
\usepackage{wrapfig}
\usepackage{amsmath, amssymb}
\usepackage{algorithm}
\usepackage{algpseudocode}

\usepackage[utf8]{inputenc}
\usepackage{tabularx} 
\usepackage{booktabs}
\usepackage{multirow}
\usepackage{graphicx}
\usepackage{array}
\usepackage{amssymb}
\usepackage{xcolor}
\usepackage{microtype}
\usepackage{amsmath}
\usepackage{inconsolata}
\usepackage{enumitem}
\usepackage{xcolor}
\definecolor{mybgcolor}{HTML}{ea998f} 
\definecolor{mybgcolor2}{HTML}{c4d1f7}
\definecolor{mybgcolor3}{HTML}{ffecc1}
\definecolor{mybgcolor4}{HTML}{aec5a8}
\definecolor{customgreen}{HTML}{A2CB32}

\newcommand{\model}{Entro-duction\xspace}
\newcommand{\modelbf}{\textbf{Entro-duction}\xspace}
\title{Entropy-based Exploration Conduction for Multi-step Reasoning}

\author{
    Jinghan Zhang\textsuperscript{\rm 1},
    Xiting Wang\textsuperscript{\rm 2}\thanks{Corresponding Author. Beijing Key Laboratory of Research on Large Models and Intelligent Governance.  Engineering Research Center of Next-Generation Intelligent Search and Recommendation, MOE.},
    Fengran Mo\textsuperscript{\rm 3}, 
    Yeyang Zhou\textsuperscript{\rm 4}, 
    Wanfu Gao\textsuperscript{\rm 5},
    Kunpeng Liu\textsuperscript{\rm 1} \\
  \textsuperscript{\rm 1}Portland State University, \\
  \textsuperscript{\rm 2}Gaoling School of Artificial Intelligence Renmin University of China Beijing, China,\\
  \textsuperscript{\rm 3}University of Montreal, 
  \textsuperscript{\rm 4}Uber,
  \textsuperscript{\rm 5}Jilin University \\
  \texttt{\{jinghanz,kunpeng\}@pdx.edu} \\
    \texttt{xitingwang@ruc.edu.cn} \\
    \texttt{fengran.mo@umontreal.ca} \\
    \texttt{yeyang.zhou@uber.com}, 
    \texttt{gaowf@jlu.edu.cn} 
}

\begin{document}
\maketitle
\begin{abstract}
Multi-step processes via large language models (LLMs) have proven effective for solving complex reasoning tasks. However, the depth of exploration of the reasoning procedure can significantly affect the task performance. 
Existing methods to automatically decide the depth often lead to high cost and a lack of flexibility. 
To address these issues, we propose \textbf{Entro}py-based Exploration Depth Con\textbf{duction} (\modelbf), a novel method that dynamically adjusts the exploration depth during multi-step reasoning by monitoring LLM's output entropy and variance entropy. 
We employ these two features to capture the model's uncertainty of the current step and the fluctuation of uncertainty across consecutive reasoning steps. 
Based on the observed entropy changes, the LLM selects whether to deepen, expand, or stop exploration according to the probability, which facilitates the trade-off between the reasoning accuracy and exploration effectiveness. 
Experimental results across four benchmark datasets demonstrate the efficacy of \model. 
\end{abstract}

\section{Introduction}
Large language models (LLMs) have demonstrated remarkable reasoning capabilities across various domains~\cite{brown2020language,touvron2023llama,patterson2022carbon,fu2022complexity,wei2022chain}. However, they would still encounter challenges in generating accurate and effective multi-step reasoning in terms of complex downstream tasks. Specifically, LLMs may exhibit over-reasoning or under-reasoning, which both imply the depth of exploration for a given problem does not match expectations~\cite{ahn2024large,mirzadeh2024gsm,huang2022towards,fu2023specializing}. This mismatch issue of reasoning path could lead to several issues: (1) inaccurate, insufficient, or redundant reasoning outcomes; (2) unnecessary computation costs~\cite{yeo2025demystifying,yang2024reinforcing,lightman2023let}. These issues might be attributed to two aspects: i) the lack of evaluation and regulatory mechanisms for LLMs' reasoning process; ii) there are significant variations in the reasoning process required for different tasks, and LLMs might not be unable to accurately judge and adjust the depth of exploration for a task solely based on their parametric knowledge during pre-training.
\begin{figure}[H]
  \centering
  \includegraphics[width=1\linewidth]{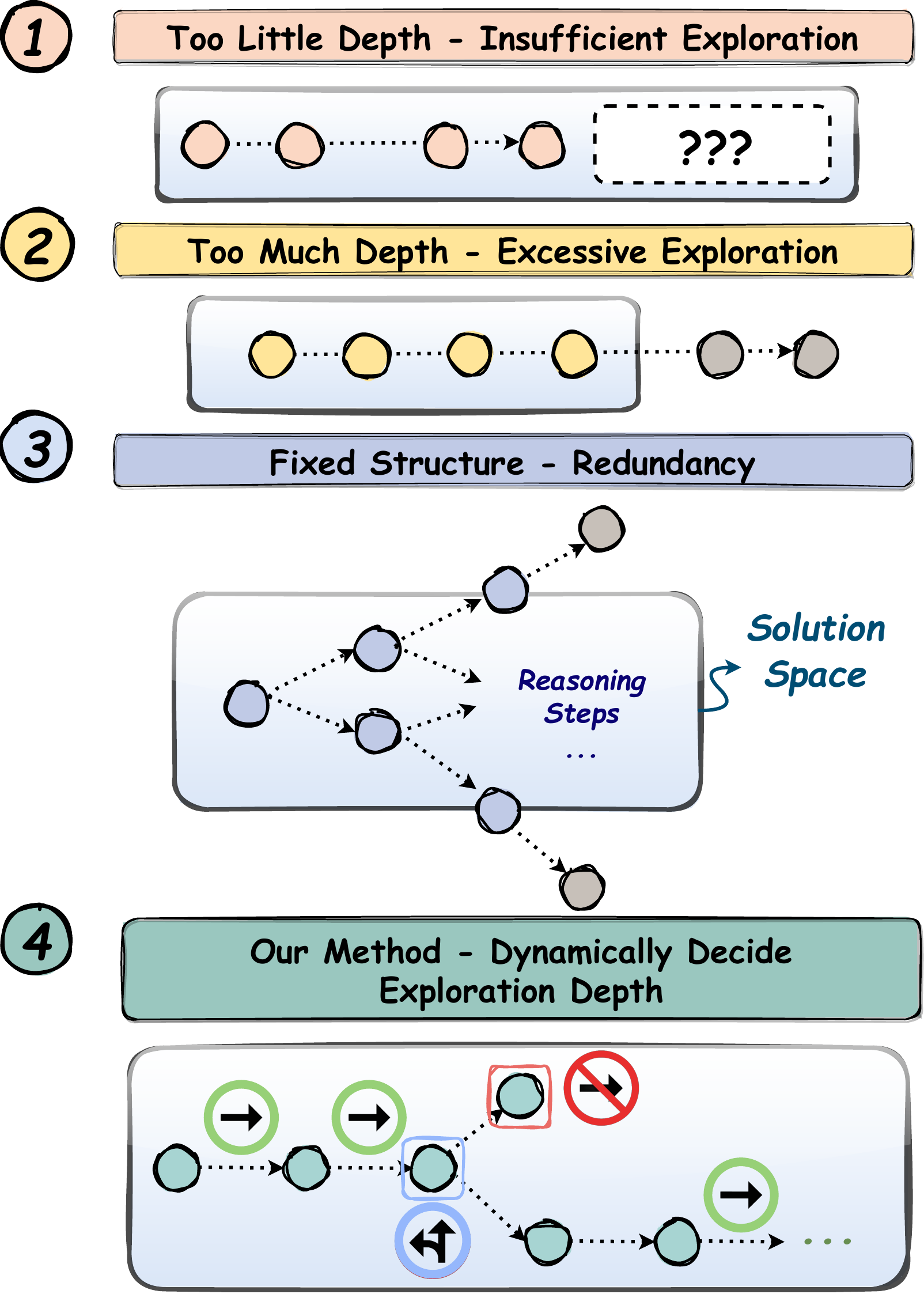}
  \caption{Reasoning depth mismatching solution space.}
  \label{fig:intro}
\end{figure}
Existing methods for optimizing the exploration depth of multi-step reasoning in LLMs can be categorized into two types: outcome-based optimization and process-based optimization. Outcome-based optimization aligns the LLMs' reasoning exploration with human expectation after generating a complete reasoning path and approaching the final conclusions~\cite{jin2024impact,liu2024languagemodelslearnskip,ton2024understandingchainofthoughtllmsinformation,yu2024distilling21}. These approaches rely on post-training techniques of LLMs, which are resource-intensive and do not provide an immediate improvement on the current task. Further, due to the diversity of reasoning tasks, the enhancement gain is task-specific. In process-based optimization, the LLMs supervise and evaluate their reasoning process through self-critique or by labeling reasoning steps to enhance outputs~\cite{ma2025cotvalvelengthcompressiblechainofthoughttuning,yang2024reinforcing,pan2023automaticallycorrectinglargelanguage}. The advantages of the process-based optimization methods are their immediacy and low cost. However, the LLMs should have high reasoning capabilities and a substantial knowledge base. Moreover, since the process is usually opaque, it is difficult for humans to effectively provide supervision signals and interpret the optimization processes. The model's illusions, biases, and errors may be reinforced during this process~\cite{stechly2023gpt4doesntknowits,stechly2024self,liang2024learning,song2025progco}.

\paragraph{Our Target.} Given the existing challenges, we aim to develop a method that guides LLMs to automatically, concisely, and transparently determine the appropriate depth of exploration based on task information and the model's reasoning state. The goal is to enable the model to look ahead during reasoning, plan dynamically, and decide whether further exploration is necessary to achieve optimal reasoning outcomes. The whole procedure involves enhancing multi-step reasoning performance and reducing unnecessary exploration.

\paragraph{Our Method.} To tackle these challenges, we propose \textbf{Entro}py-based Exploration Depth Con\textbf{duction} (\modelbf), a novel method to help LLMs assess the adequacy of exploration during multi-step reasoning processes, thus enhancing the outcomes of reasoning. Inspired by Entropy Uncertainty Measurement~\cite{coles2017entropic,farquhar2024detecting,zhang2024prototypical,rosenfeld1996maximum}, we employ entropy changes in the LLM's reasoning process to evaluate its uncertainty of reasoning, which reflects the reasoning confidence, and accordingly switch exploration strategy. Specifically, we use entropy and variance entropy as rewards to update the LLM’s probability distribution for its next exploratory action, whether to deepen, stop, or expand exploration. This distribution subsequently guides a behavior selection mechanism that promotes reasoning when exploration is insufficient and avoids redundant reasoning when it is adequate.

In summary, our contributions involve:
\begin{enumerate}
 \item We propose \model to help LLMs dynamically evaluate the adequacy of exploration based on their reasoning uncertainty to enhance reasoning performance and avoid unnecessary exploration.
 \item We further design an entropy-based exploration behavior selection mechanism, which refers to LLMs' expectancy and confidence in the reasoning procedure.
 \item We conduct a series of experiments to demonstrate the effectiveness of \model on various reasoning tasks. Our results and analysis show that the \model outperforms baseline methods.
\end{enumerate}
\section{Related Work}

\paragraph{Reasoning Steps and Structures.} When responding to queries, LLMs typically provide direct outputs. For complex questions, direct outputs often fail to deliver correct answers because they may overlook deeper logical connections or contextual information~\cite{xia2024beyond,minaee2024large}. Multi-step reasoning involves instructing LLMs to decompose and progressively address problems, breaking down complex tasks into smaller, manageable units to significantly enhance reasoning capabilities~\cite {chu2023survey}.
The simplest structure of multi-step reasoning is the Chain of Thought (CoT)~\cite{wei2022chain,wang2024chain}, which links reasoning steps by connecting distinct thoughts into a linear, coherent sequence~\cite{li2024chain,jin2024self,sprague2024cot}. 

To enable more comprehensive exploratory reasoning, some studies based on CoT have developed structured reasoning methods, such as self-consistent CoT (CoT-SC), Complex CoT, and Tree-of-Thought (ToT)~\cite{wang2022self,zhang2024diagram,yao2024tree,mo2024survey,liu2024logic,mo2024tree,zhang2024ratt}. 
These methods are called reasoning structures. They guide LLMs to do multi-directional exploration of problem solution spaces for superior reasoning solutions by capturing more complex and varied logical relationships~\cite{xia2024beyond,stechly2024chain,mo2023convgqr,liang2024internal}. However, the breadth and depth of these reasoning structures highly depend on predefined settings and vary greatly across different tasks, limiting their generalizability and flexibility.

\paragraph{Optimization of Reasoning Depth.} The depth of reasoning structures refers to the number of layers or steps in the reasoning process, namely, the number of reasoning steps an LLM must undertake before reaching a final answer~\cite{plaat2024reasoning,TreeOfThoughts}. For any given task, the optimal number of reasoning layers often correlates with the task's complexity and the level of detail required~\cite{zhang2024diagram}. Current methods in determining these layers or optimize reasoning structures automatically. These methods include using reinforcement learning algorithms to optimize the number of reasoning steps or dynamically adjusting the depth of exploration during the reasoning process~\cite{jin2024impact,liu2024languagemodelslearnskip,hoffmann2022training}.

The main issues of these methods include: (1) the automated algorithms may lack precision due to the lack of precision; (2) making dynamic adjustments without fully understanding task characteristics could harm the reasoning process; (3) for highly complex or novel tasks, preset reasoning structures may be inappropriate, and could limit the model's applicability and flexibility. These issues together ultimately affect the LLM's reasoning reliability and efficiency, and lead to inaccurate reasoning outcomes or redundant exploration.
\section{Methodology}

\subsection{Problem Formulation}


Given a reasoning task and an LLM $\mathcal{R}$ as the reasoner, the multi-step reasoning process is to generate a reasoning structure $\mathcal{S}$. Structure $\mathcal{S}$ comprises directed links connecting sentence-level reasoning nodes. A reasoning chain is a unidirectional sequence that begins at an initial reasoning node and concludes at a terminal node:
\begin{equation}
    \mathcal{C}_i= \mathcal{T}_{i1} \rightarrow \mathcal{T}_{i2} \rightarrow \cdots \rightarrow \mathcal{T}_{i,j} \rightarrow \cdots, i=1,...,m,
\end{equation}

\noindent where $\mathcal{T}_{i,j}$ is the $j$-th node in the $i$-th chain,$m$ is the total number of chains in $\mathcal{S}$. The chain length $|\mathcal{C}_i|$ is the total number of nodes within $\mathcal{C}_i$.
We define the depth $\mathcal{S}_d $ of $\mathcal{S}$ as the maximum length of any reasoning chain within the structure:
\begin{equation}
\mathcal{S}_d = \max\{|\mathcal{C}_i|\}.
\end{equation}

Since $\mathcal{S}$ could contain several valid reasoning chains, the reasoning conclusion $L$ is made through a voting mechanism $V$:
\begin{equation}
L = V(\mathcal{S}, \mathcal{R}).
\end{equation}
In this paper, our goal is to generate a structure $\mathcal{S}$ such that it achieves optimal reasoning accuracy, denoted as $\text{Acc}(L)$, with an optimal depth $\mathcal{S}_d$.

\subsection{Reasoning State Evaluation}
\begin{figure*}
    \centering
    \includegraphics[width=0.95\linewidth]{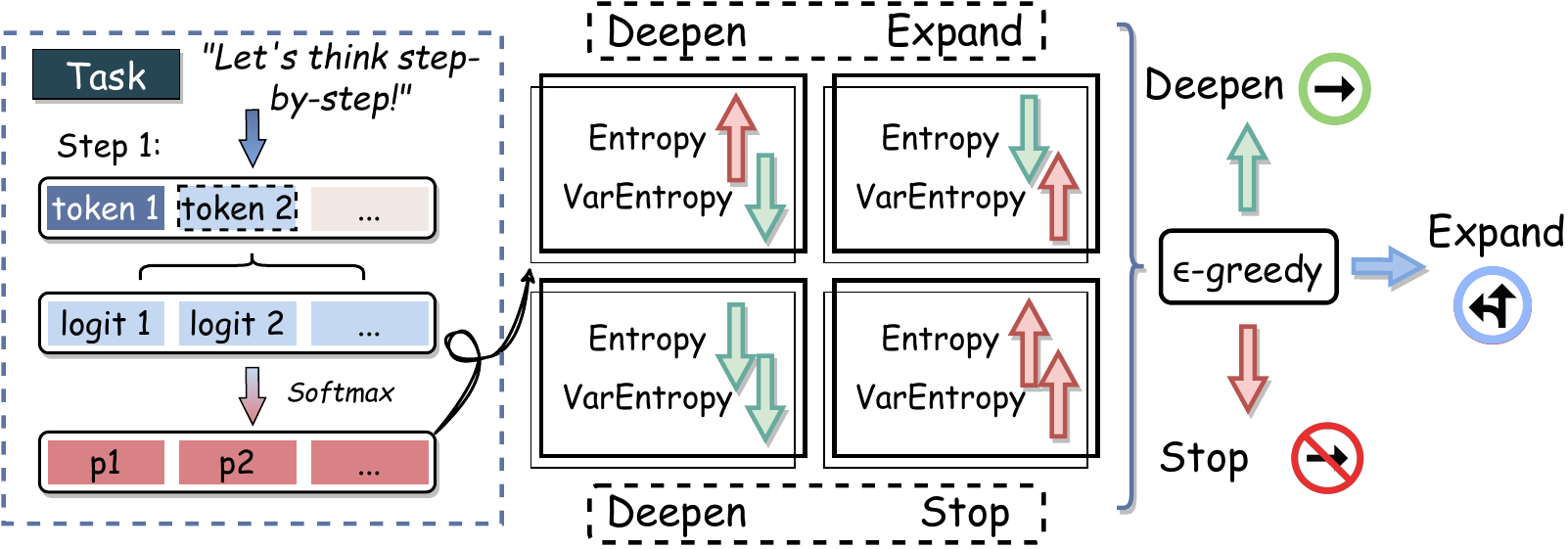}
    \caption{Framework of \model. We obtain two metrics, entropy and variance entropy, by calculating the probabilities of the logits at each reasoning step. Subsequently, we employ the $epsilon$-greedy method to select the appropriate exploration behavior based on changes in both metrics.}
    \label{fig:method}
    \vspace{-2mm}
\end{figure*}
The essence of the multi-step reasoning process is to explore the solution space of task $\mathcal{Q}$. Our exploration goal is to cover as many potential reasoning paths as possible to ensure the accuracy and completeness of the solution. However, exhaustive exploration is inefficient and often impractical. As the problem’s complexity grows, the solution space expands exponentially, driving the computational and time costs to untenable levels. Consequently, we must balance the breadth and depth of exploration. Achieving this balance calls for a method that can look forward from each current reasoning state, predicting and adjusting subsequent exploration steps, so that we do not miss crucial paths or waste resources on unnecessary ones.

We employ uncertainty and stability to describe the reasoning state. Uncertainty measures the divergence of current thought processes, as shown in Figure~\ref{fig:method}. In a reasoning step, a high uncertainty indicates the presence of multiple possible directions or conclusions, this means a wide scope of exploration is necessary. Conversely, low uncertainty, where there are few or even a single possible outcome, indicates a more focused path. Specifically, we employ entropy as a metric for uncertainty to quantify the number of potential paths that need exploration at any given moment and to gauge the confidence level in the conclusions.

\textbf{\textit{Definition 1: Entropy.}} Consider a reasoning step represented by a sentence, denoted as $\mathcal{T}_{i,j}$, which consists of a sequence of $n$ tokens:

\begin{equation}
\mathcal{T}_{i,j} = \{ t_{ij1}, t_{ij2}, \ldots, t_{ijn} \}.
\end{equation}
Each token $t_{ijk}$ matches a logit $l_{ijk}$, which is the model’s raw output before the softmax function. The collection of logits for the entire sentence is:
\begin{equation}
l_{ij} = \{ l_{ij1}, l_{ij2}, \ldots, l_{ijn} \}.
\end{equation}

We calculate the probability $p_{ijk}$ of each token $t_{ijk}$  by applying the softmax function to its corresponding logit $l_{ijk}$:
\begin{equation}
p_{ijk} = \frac{\exp(l_{ijk})}{\sum_{r=1}^n \exp(l_{ijr})}.
\end{equation}

\noindent Then the entropy of the sentence $\mathcal{T}_{i,j}$ is:
\begin{equation}
H(\mathcal{T}_{i,j}) = -\sum_{k=1}^n p_{ijk} \log_2(p_{ijk}).
\label{eq:entropy}
\end{equation}
This measures the uncertainty or information content encoded in the probability distribution $\{p_{ij1}, p_{ij2}, \ldots, p_{ijn}\}$.

To compare the entropies across reasoning steps of varying lengths, we define the normalized entropy as:
\begin{equation}
\widetilde{H}(\mathcal{T}_{i,j}) = \frac{H(\mathcal{T}_{i,j})}{\log_2(n)}.
\end{equation}
Here, $\log_2(n)$ is the maximum possible entropy when all $n$ tokens have uniform probability. Hence, the normalized entropy is bounded between 0 and 1 for consistent comparisons.

Similarly, we employ variance entropy to capture how much uncertainty fluctuates across consecutive reasoning steps. It indicates the consistency or divergence of the thought process.

\textbf{\textit{Definition 2: Variance Entropy.}} For reasoning step $\mathcal{T}_{i,j}$ of length $n$, let:
\begin{equation}
\overline{H}(\mathcal{T}_{i,j}) = \frac{1}{n} \sum_{k=1}^n H(t_{ijk}),
\end{equation}
be the average token-level entropy in $\mathcal{T}_{i,j}$. We define the variance entropy as:
\begin{equation}
\sigma^2_H(\mathcal{T}_{i,j}) = \frac{1}{n} \sum_{k=1}^n \left[H(t_{ijk}) - \overline{H}(\mathcal{T}_{i,j})\right]^2.
\label{eq:var}
\end{equation}
For comparisons, we define the normalized variance entropy:
\begin{equation}
\widetilde{\sigma^2_H}(\mathcal{T}_{i,j}) = \frac{\sigma^2_H(\mathcal{T}_{i,j})}{\log_2(n)}.
\end{equation}
In this way, we have a normalized concise metric for tracking fluctuations in uncertainty within each reasoning step.

\subsection{Exploration Behaviors}

With two metrics for reasoning state defined above, we further consider how to use changes in these metrics to determine exploration behavior strategies. The possible scenarios are listed below:
\begin{enumerate}

    \item \textbf{\textit{Entropy \textcolor{customgreen}{$\downarrow$} , Variance Entropy \textcolor{customgreen}{$\downarrow$}}:} The reasoning step becomes more certain, and the overall thought process more coherent. This indicates that information is becoming more focused, and the reasoning process is stable and effective. The LLM should continue to explore in this direction.
    \item \textbf{\textit{Entropy \textcolor{red}{$\uparrow$} , Variance Entropy \textcolor{customgreen}{$\downarrow$}}:} \label{item:entropy-variance-behavior} The reasoning step introduces more uncertainty, but the fluctuations between different steps are decreasing. This suggests that while a broader range of possibilities has emerged, the overall direction has not become dispersed. The LLM should continue to explore in this direction.
    \item \textbf{\textit{Entropy \textcolor{customgreen}{$\downarrow$} , Variance Entropy \textcolor{red}{$\uparrow$}}:} \label{item:entropy-variance-behavior2} The uncertainty of reasoning is decreasing, but the fluctuation between reasoning steps is increasing. This indicates potential divergences in local steps, and we should consider increasing exploration in different directions to cover possible solutions.
    \item \textbf{\textit{Entropy \textcolor{red}{$\uparrow$} , Variance Entropy \textcolor{red}{$\uparrow$}}:} The reasoning process becomes simultaneously more complex and more unstable. This indicates that current exploration might have strategic deviations, but another possibility is that the model is exploring a new or more challenging direction. We need to consider avoiding ineffective exploration while maintaining the potential for the model to tackle challenging problems.
\end{enumerate}

Accordingly, we design a mechanism that adjusts the probability of exploration behaviors based on entropy and variance entropy changes. We define the exploration behaviors as follows:

\textbf{\textit{Deepen:}} This behavior extends the current reasoning chain $\mathcal{C}_i$ by adding a new node $ \mathcal{T}_{i,j+1}$:
\begin{equation}
\mathcal{C}_i \rightarrow \mathcal{C}_i \cup \{\mathcal{T}_{i,j+1}\}.
\end{equation}

\textbf{\textit{Expand:}} This behavior divides the current reasoning chain at $\mathcal{T}_{i,j} $, creates two separate chains $\mathcal{C}_i$ and $\mathcal{C'}_i$. Each chain extending from the split point generates a new node:
\begin{equation}
\mathcal{C}_i \rightarrow \left( \mathcal{C}_i \cup \{\mathcal{T}_{i,j+1}\} \right), 
\mathcal{C'}_i \rightarrow \left(\mathcal{C}_i \cup \{\mathcal{T'}_{i,j+1}\} \right).
\end{equation}

\textbf{\textit{Stop:}} This behavior terminates the extension of the current chain $\mathcal{C}_i$ at the current node $\mathcal{T}_{i,j}$:
\begin{equation}
\mathcal{C}_i \rightarrow \mathcal{C}_i \setminus \{\mathcal{T}_{i,j+1}\}.
\end{equation}


\subsection{Behavior Selection Mechanism}

At the $j$-th reasoning step, we define:
\begin{equation}
    \Delta H_j = H(\mathcal{T}_{j+1}) - H(\mathcal{T}_{j}),
\label{eq:dte}
\end{equation}
\begin{equation}
    \Delta \sigma_{H,j}^2 = \sigma_{H}^2(\mathcal{T}_{j+1}) - \sigma_{H}^2(\mathcal{T}_{j}),
\label{eq:dtv}
\end{equation}
\noindent which denotes the changes in entropy and variance entropy, respectively. We define the state:
\begin{equation}
\mathbf{s}_j = \bigl(\Delta H_j,\Delta \sigma_{H,j}^2\bigr),
\end{equation}
and the set of possible actions as 
\begin{equation}
\mathcal{A} = \{\mathrm{Deepen},\mathrm{Expand},\mathrm{Stop}\}.
\end{equation}

We introduce a mapping function $\Phi:\mathbb{R}^2 \rightarrow \mathcal{A}$
that assigns to each state $(\Delta H_j,\,\Delta \sigma_{H,j}^2)$ a ``best'' action $a_j^*$:

{\small
\begin{equation}
\Phi(\Delta H, \Delta \sigma_H^2) 
= \begin{cases}
\mathrm{Deepen}, & \begin{array}{@{}l@{}}\text{if } (\Delta H < 0, \Delta \sigma_H^2 < 0) \\ \text{or } (\Delta H > 0, \Delta \sigma_H^2 < 0),\end{array} \\[8pt]
\mathrm{Expand}, & \text{if } \Delta H < 0, \Delta \sigma_H^2 > 0, \\[4pt]
\mathrm{Stop}, & \text{if } \Delta H > 0, \Delta \sigma_H^2 > 0.
\end{cases}
\label{eq:phi}
\end{equation}}
Then, at each step $j$, we sample the actual action $a_j$ according to an $\epsilon$-greedy rule:
\begin{equation}
\pi_j\bigl(a\mid \mathbf{s}_j\bigr)
=\,
\begin{cases}
1-\epsilon, & a = a_j^*,\\[4pt]
\dfrac{\epsilon}{|\mathcal{A}|\,-\,1}, & a \neq a_j^*.
\end{cases}
\label{eq:pi}
\end{equation}

Given the current state $\mathbf{s}_j=(\Delta H_j, \Delta \sigma_{H,j}^2)$, we first compute $a_j^*=\Phi(\mathbf{s}_j)$, then draw $a_j$ from $\pi_j(\cdot\mid \mathbf{s}_j)$. If $a_j=$ \textbf{\textit{Stop}}, the reasoning ends; otherwise, the system transitions to the next state $\mathbf{s}_{j+1}$, where the new entropy measures yield an updated state.

\begin{algorithm}[t]
\caption{\modelbf}
\label{alg:entro-duction-slim}
\begin{algorithmic}[1]
\State \textbf{Input}:
  Reasoning task $\mathcal{Q}$; LLM reasoner $\mathcal{R}$; 
  max steps $J$; exploration rate $\epsilon$.
\State \textbf{Output}:
  Reasoning structure $\mathcal{S}$ and conclusion $L$.

\State Initialize $\mathcal{S}$; set $j \gets 1$; initialize chains $\{\mathcal{C}_i\}$.

\While{$j \leq J$}
  \For{each active chain $\mathcal{C}_i$}
    \State Compute $H(T_j)$ and $\sigma_H^2(\mathcal{T}_j)$ according to Eqs.~\ref{eq:entropy},~\ref{eq:var}.
    \State Compute $\Delta H_j$ and $\Delta \sigma_{H,j}^2$ according to Eqs.~\ref{eq:dte},~\ref{eq:dtv}).
    \State Determine $a_j^* \gets \Phi(\Delta H_j, \Delta \sigma_{H,j}^2)$ according to Eq.~\ref{eq:phi}.
    \State Sample action $a_j$ with probability $\pi_j(a\mid s_j)$ according to Eq.~~\ref{eq:pi}).
    \State \textbf{Execute} $a_j$:
      \begin{enumerate}[label=\roman*)]
        \item \textbf{Deepen}: append $\mathcal{T}_{j+1}$ to $\mathcal{C}_i$.
        \item \textbf{Expand}: branch $\mathcal{C}_i$ into two chains with $\mathcal{T}_{j+1}$ and $\mathcal{T}'_{j+1}$.
        \item \textbf{Stop}: finalize $\mathcal{C}_i$ (no further expansion).
      \end{enumerate}
  \EndFor
  \If{all chains stopped or $j = J$}
    \State \textbf{break}
  \EndIf
  \State $j \gets j + 1$
\EndWhile

\State $L \gets V(\mathcal{S}, \mathcal{R})$ (final conclusion via consensus).
\State \textbf{return} $\mathcal{S}, \; L$
\end{algorithmic}
\end{algorithm}

\section{Experiments}
In this section, we first compare the reasoning performance and reasoning steps used between baseline methods and \model. Subsequently, we present ablation studies to analyze the contributions of each part of our strategies. Following this, we examine how different parameter settings impact \model's overall robustness.
\subsection{Experiment Settings}
\paragraph{Datasets.} \model is a general approach applicable to various LLMs and reasoning tasks. Here, we test across four reasoning tasks with benchmark datasets, including two mathematical tasks (GSM8K~\cite{cobbe2021gsm8k}, SVAMP~\cite{patel2021nlp}) and two commonsense question-answering tasks (StrategyQA~\cite{geva2021did}, CommonsenseQA~\cite{talmor2018commonsenseqa}). Here, GSM8K challenges language models with multi-step math reasoning tasks, assessing their complex reasoning capabilities, while SVAMP focuses on simpler, one-step math reasoning tasks. StrategyQA tests strategic reasoning skills for deriving implicit strategies and using deductive reasoning to answer questions. CommonsenseQA (CSQA) tests the ability to handle commonsense reasoning with everyday knowledge. In evaluating performance on these datasets, we primarily focus on reasoning accuracy (\%) as the key metric.
\paragraph{Baselines.} We compare \model with two strong baseline types: (1) Reasoning structures, including Chain of Thought (CoT), Chain of Thought with Self-Consistency (CoT-SC), Tree of Thought (ToT) and Complex CoT:
\begin{itemize}
    \item CoT: Guides the model to solve problems step-by-step and generates a coherent reasoning chain that leads to a conclusion.
    \item CoT-SC: Generates multiple reasoning chains and uses a majority vote to determine the final output. We sample answer $8$ (CoT-SC@maj8) and $64$ (CoT-SC@maj64 ) times to employ majority vote for selection.
    \item ToT: Expands the reasoning process into a tree-like structure where multiple branches represent different reasoning pathways.
    \item Complex CoT: Engages with complex samples and selects the best solution from various intricate reasoning paths for tackling multi-faceted and challenging problems.
\end{itemize}
and (2) Reasoning depth optimization methods, including Self-talk~\cite{shwartz2020unsupervised,molfese2024zebra} and Distillation-Reinforcement-Reasoning (DRR)~\cite{yang2024reinforcing}:
\begin{itemize}
    \item Self-talk: Enhances reasoning by eliciting LLMs to generate exploratory questions, uncovering implicit background knowledge and selecting the best answer.
    \item DRR: Distills LLM reasoning processes into synthetic data by training a lightweight model to provide feedback.

\end{itemize}
For detailed settings of baselines, please refer to Appendix~\ref{sec:baselines}.

\paragraph{Implementation Details.} We conduct the experiments utilizing the \texttt{Llama-3.1-8B-Instruct}\footnote{\url{https://huggingface.co/meta-llama/Llama-3.1-8B-Instruct}}. The temperature for all models is set to the default value of 0.7, with a maximum token limit of 128. All tasks are performed on an NVIDIA 4090 GPU.
\subsection{Overall Performance}
The overall performance is reported in Table~\ref{tab:overall}. 
\begin{table*}[htbp]
    \centering
    \resizebox{\textwidth}{!}{ 
    \begin{tabular}{l|cc|cc|cc|cc}
        \toprule
        \hline
        \multirow{3}{*}{\textbf{Method}} & \multicolumn{4}{c|}{\textbf{Math}} & \multicolumn{4}{c}{\textbf{Commonsense}} \\
        \cline{2-9}
        & \multicolumn{2}{c|}{GSM8K} & \multicolumn{2}{c|}{SVAMP} & \multicolumn{2}{c|}{StrategyQA} & \multicolumn{2}{c}{CSQA} \\
        \cline{2-9}
        & Accuracy & \# Steps & Accuracy & \# Steps & Accuracy & \# Steps & Accuracy & \# Steps \\
        \midrule
        CoT & 75.2 & 8.0 & 83.4 & 8.0 & 57.7 & 5.0 & 75.6 & 5.0 \\
        CoT-SC@maj8 & 78.1 & 24.0 & 87.5 & 24.0 & 68.3 & 15.0 & 78.2 & 15.0 \\
        CoT-SC@maj64 & 80.2 & 24.0 & 89.6 & 24.0 & 67.1 & 15.0 & 78.7 & 15.0 \\
        ToT & 72.6 & 121.0 & 83.3 & 121.0 & 65.8 & 121.0 & 73.5 & 121.0 \\
        Complex CoT & 81.4 & 8.0 & 86.2 & 8.0 & 65.7 & 5.0 & 73.9 & 5.0 \\
        \midrule
        Self-talk & 79.1 & / & 83.7 & / & 61.5 & / & 70.0 & / \\
        DRR & 83.0 & / & 90.2 & / & 67.7 & / & 82.1 & / \\
        \midrule
        \textbf{Entro-duction} & 85.4 & 9.5 & 92.0 & 11.20 & 70.3 & 9.6 & 79.6 & 7.1 \\
        \hline
        \bottomrule
    \end{tabular}
    }
    \caption{Performance comparison across different reasoning methods with accuracy and number of steps.}
    \label{tab:overall}
\end{table*}
For the baselines, we compare reasoning accuracies across four datasets, and for reasoning structures, we additionally measure the number of steps required. Since each structure’s steps and branches are predefined, we adopt configurations that can perform well and that further increasing the step count often does not bring significant gains. Specifically, for math tasks, CoT is set to 8 steps, CoT-SC adopts 3 parallel chains of 8 steps each, and ToT generates three branches per step for five layers. For commonsense tasks, CoT is set to 5 steps, CoT-SC adopts 3 parallel chains of 5 steps each, and ToT generates three branches per step for five layers.

Compared to reasoning structures, our \model approach achieves both higher accuracy and fewer reasoning steps. For instance, on GSM8K, CoT reaches 0.75 accuracy, CoT-SC@maj64 0.80, and Complex CoT 0.81, while \model attains 0.85. A similar advantage appears on SVAMP (up to 0.92), and on the commonsense tasks StrategyQA and CSQA, \model scores 0.70 and 0.79 respectively, surpassing most baselines. Moreover, tree-structured methods ToT require hundreds of steps (over 100 in several cases), while \model needs much fewer steps, only more than CoT. Even compared to fixed-step approaches such as CoT and Complex CoT, \model delivers higher accuracy in a similar or slightly increased number of steps.

Compared to reasoning depth optimization methods, \model consistently attains higher performance. Self-talk achieves accuracies of 0.79, 0.61, and 0.70 on GSM8K, StrategyQA, and CSQA, all below \model’s 0.85, 0.70, and 0.79. DRR demonstrates decent performance on SVAMP (0.90) and CSQA (0.82), but still trails \model on GSM8K (0.83 vs. 0.85) and StrategyQA (0.67 vs. 0.70). Moreover, these methods often rely on additional training or separate models, while \model balances accuracy and a relatively low reasoning overhead without training.

\subsection{Ablation Study}
\subsubsection{Impact of Jointly Using Entropy and Variance Entropy} 
To validate the necessity of jointly using entropy and variance entropy, we conduct experiments across four datasets to validate the necessity of jointly using entropy and variance entropy. We set four different scenarios: \textit{Base} (neither used), \textit{Entropy} (only entropy used), \textit{Variance} (only variance entropy used), and \textit{Both} (both used).

\begin{figure}
  \centering
  \includegraphics[width=1\linewidth]{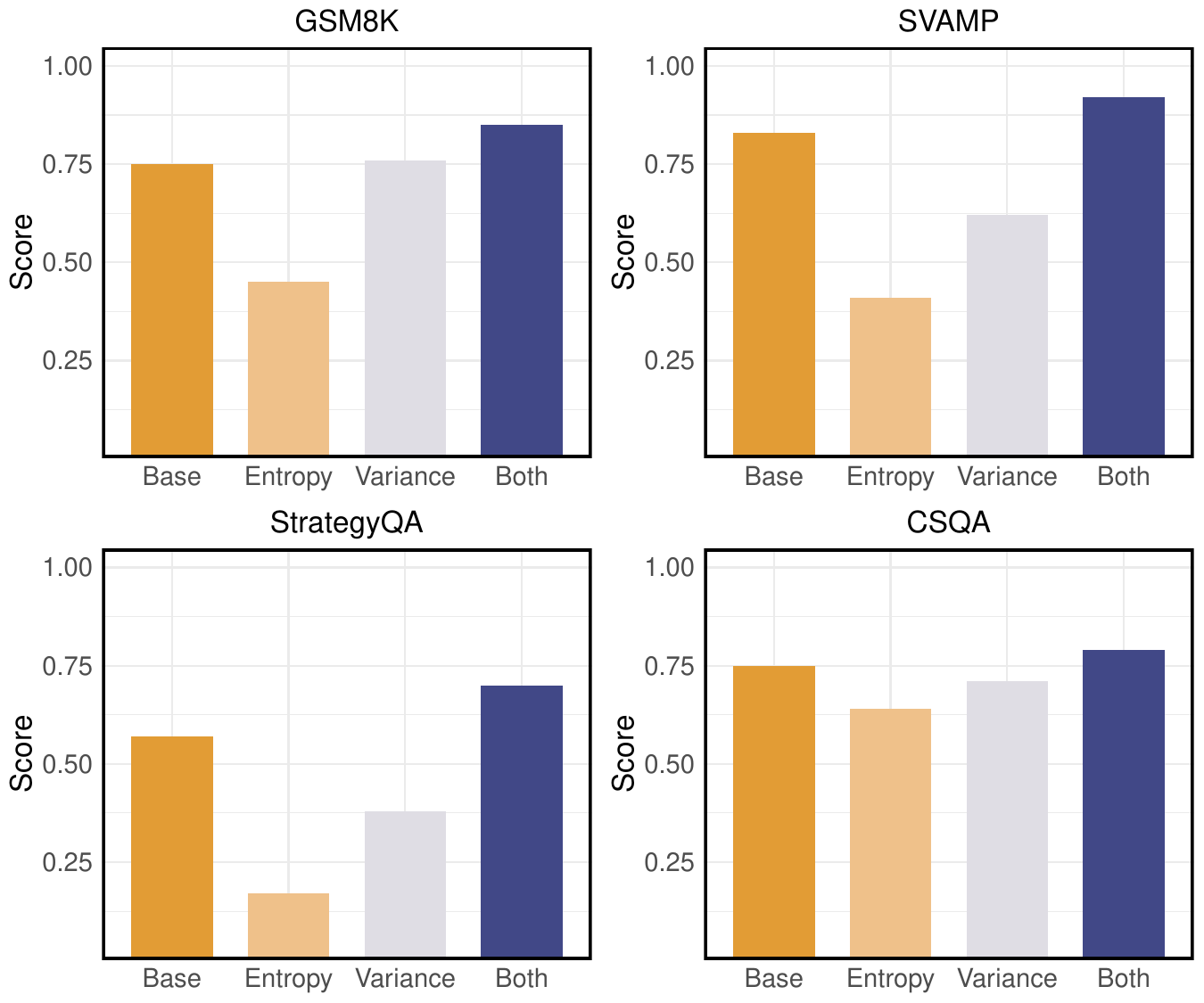}
  \caption{Comparison of adjusting with entropy and/or variance entropy.}
  \label{fig:aba1}
  \vspace{-3mm}
\end{figure}
As shown in Figure~\ref{fig:aba1}, when using only entropy, the model tends to stop reasoning prematurely in scenarios with many potential outcomes but fewer overall fluctuations (Scenario~\ref{item:entropy-variance-behavior}). Using only variance entropy can capture changes in fluctuations between reasoning steps. It slightly outperforms using entropy alone, but still proves inadequate for handling various uncertainty scenarios Scenario~\ref{item:entropy-variance-behavior2}), with accuracies mostly close to or below \textit{Base}.
\subsubsection{Impact of Expansion in Reasoning}
Compared with \textit{\textbf{Deepen}} and \textit{\textbf{Stop}} that directly affect reasoning depth, \textit{\textbf{Expand}} is a key behavior in \model to branch out the current reasoning path to cover more potential solutions. We further validate the necessity of the behavior \textit{\textbf{Expand}} by comparing three settings across datasets: \textit{Base} (no behavior selection), \textit{w/o} (only using \textit{\textbf{Deepen}} and \textit{\textbf{Stop}}), and \textit{w/} (enabling \textit{\textbf{Expand}}).
\begin{figure}
  \centering
  \vspace{-2mm}
  \includegraphics[width=1\linewidth]{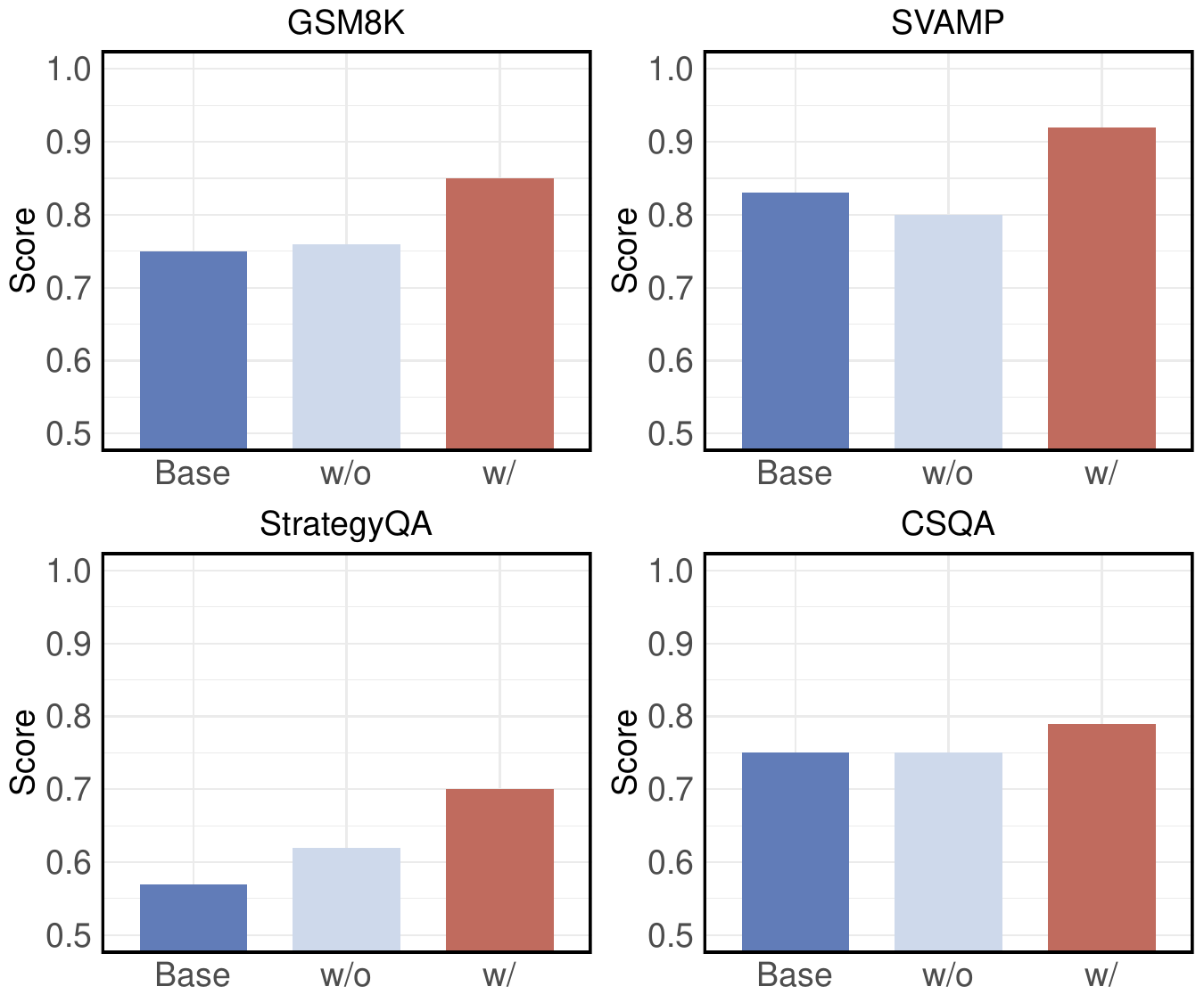}
  \caption{Impact of the behavior \textbf{\textit{Expand}}.}
  \label{fig:aba2}
  \vspace{-3mm}
\end{figure}

As shown in Figure~\ref{fig:aba2}, the accuracies of \textit{w/o} are generally lower than those of \textit{w/}, particularly in tasks requiring multiple reasoning paths or branching thought processes, such as SVAMP and StrategyQA. The result indicates that using only \textit{\textbf{Deepen}} and \textit{\textbf{Stop}} limits the exploration of potential directions, while expanding the exploration contributes to improving the completeness of the reasoning. 

\subsubsection{Impact of Soft Stop}
In some complex tasks, we notice that the initial reasoning process could see an increase in both entropy and variance entropy. However, the model may still expect valid exploration in the continued reasoning. In this case, if we adopt a ``hard stop'', which means immediate shutdown, it could terminate exploration in advance of arriving at the correct conclusion. Instead, we introduce a ``soft stop'' mechanism to balance the need for thorough exploration and the risk of redundant reasoning. The model continues for several additional steps before stopping. In our experiments, we implement four settings: \textit{Base} (no stopping strategy), \textit{Stop@1} (hard stop with immediate termination), \textit{Stop@2} (soft stop with one more reasoning step before stopping) and \textit{Stop@3} (soft stop with two more reasoning steps before stopping).
\begin{figure}[h]
  \centering
  \vspace{-2mm}
  \includegraphics[width=1\linewidth]{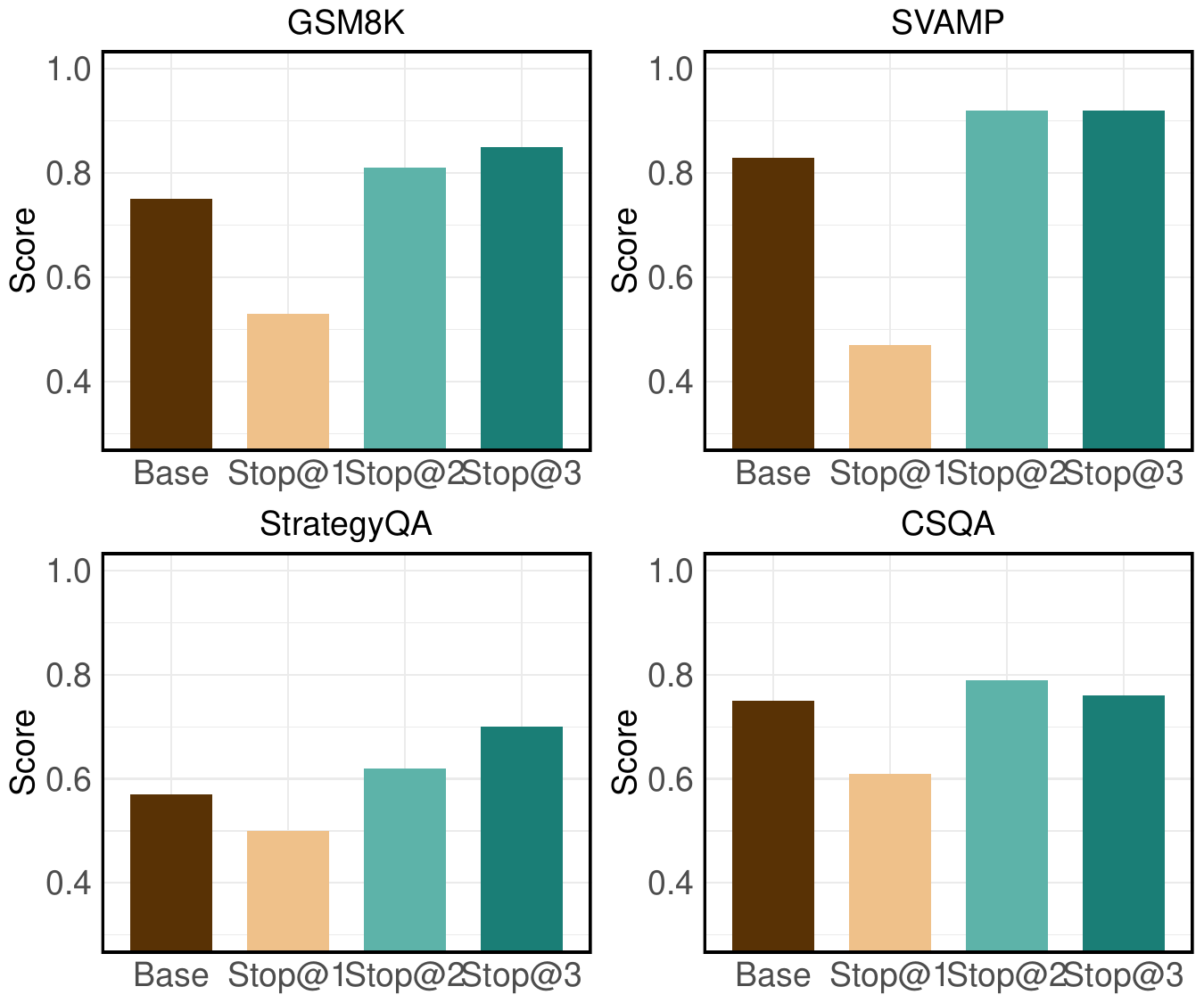}
  \caption{Comparison of stopping strategies.}
  \label{fig:aba3}

\end{figure}

As shown in Figure~\ref{fig:aba3}, we can see that the hard stop strategy has the lowest outcomes across all four datasets, lower than not employing any stopping strategy. The soft stop strategy consistently has the best results. Moreover, on datasets SVAMP and CSQA, Stop@2 performs as well or better than Stop@3. This result suggests that a soft stop extending two to three steps is sufficient to complete effective exploration without the need for extensive further reasoning.

\subsection{Robustness Study}
We further discuss the impact of $\epsilon$-greedy strategy by testing four $\epsilon$ values (0.05, 0.1, 0.25, 0.5).
\begin{figure}[h]
  \centering
  \vspace{-2mm}
  \includegraphics[width=1\linewidth]{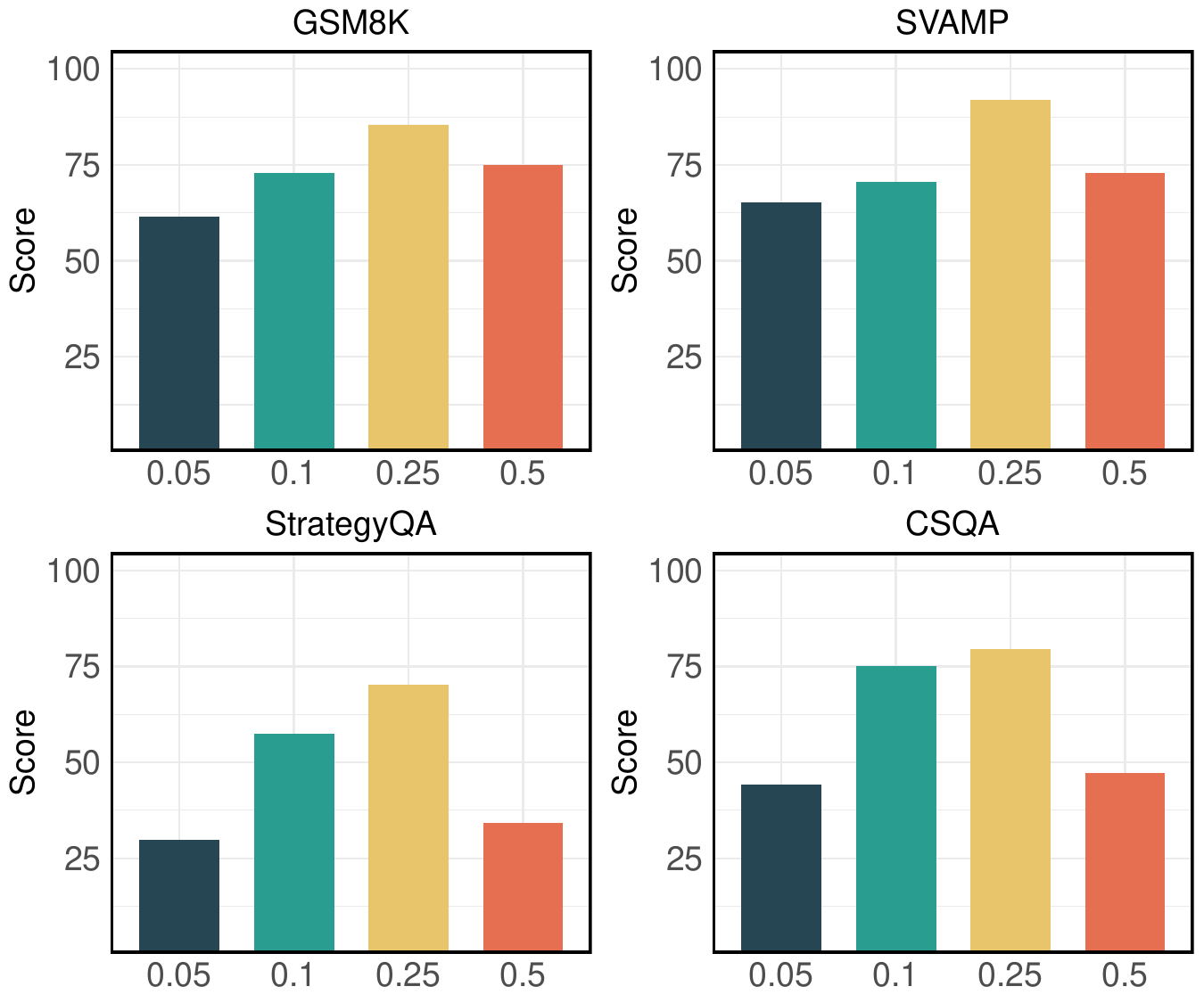}
  \caption{Comparison of choice of $\epsilon$.}
  \label{fig:epsi}
  \vspace{-2mm}
\end{figure}

As shown in Figure~\ref{fig:epsi}, $\epsilon=0.25$ consistently achieves the highest accuracy across all four datasets. This result demonstrates that this value enhances the model's performance by effectively exploring the solution space. Specifically, when $\epsilon$ is set too low ($\epsilon=0.05$), the model's performance is poor. This is likely due to insufficient exploration that relies heavily on the known strategy, thus unable to explore potential solutions hidden in the space. Meanwhile, when $\epsilon=0.5$, it outperforms $\epsilon=0.1$ in mathematical tasks and underperforms in commonsense tasks. This result indicates that tasks requiring reasoning with stringent logical structure and relatively more steps need broader exploration to identify the correct solutions. For commonsense tasks, which require more precise adopting of knowledge for quick decision-making. In this type of task, over-exploration may lead the model away from the question background and thus miss the intuitive commonsense answers.
\section{Conclusion}
In this study, we introduce \model, a novel approach that dynamically adjusts the exploration depth during LLM multi-step reasoning by monitoring the entropy and variance entropy. \model leverages the change of both metrics to select exploration behavior to enhance reasoning performance and avoid redundant reasoning steps. Our experiments across multiple reasoning datasets demonstrate the effectiveness of the \model and its components.

\newpage
\section*{Limitations}
We develop a framework with dynamic depth adjustment strategies for LLMs. If not precisely calibrated, it might lead to suboptimal reasoning performance. 
This may limit the \model method's ability to adaptively balance exploration and exploitation in real-time. Besides, the experiments are conducted on four benchmark datasets on one Llama model, which may not provide a comprehensive view of the \model's generalization capability across LLMs with varying sizes and pre-training processes. 
Moreover, the \model is mainly evaluated on specific tasks and these tasks cannot fully reflect the complexities of real-world scenarios where reasoning tasks can be variable and with more complex and external solution spaces. We left these potential explorations as future work.

\paragraph{Acknowledgements}
Dr. Xiting Wang is supported by the National Natural Science Foundation of China (NSFC) (NO. 62476279, NO. 92470205), Major Innovation \& Planning Interdisciplinary Platform for the ``Double-First Class'' Initiative, Renmin University of China, the Fundamental Research Funds for the Central Universities, and the Research Funds of Renmin University of China No. 24XNKJ18.
Dr. Xiting Wang is supported by fund for building world-class universities (disciplines) of Renmin University of China
and Public Computing Cloud, Renmin University of China.

\bibliography{ref}

\clearpage
\appendix
\section*{Appendix}
\section{Experimental Settings}
\label{sec:baselines}
\subsection{Baselines}
In CoT, we prompt the model with the sentense \texttt{``Let's think step-by-step.''}. For CoT-SC, we use a majority vote to identify the most probable correct solution, with the same few-shot examples as the standard CoT method. For Complex CoT we follow the setting in \cite{zhou2022least}. The setting of Self-talk and DRR method follows the setting in \cite{yang2024reinforcing}.

\subsection{Answer-Cleaning}

We showcase our answer-cleaning process with GSM8K as an example. 
\begin{algorithm}
\caption{Answer Cleansing for GSM8K Dataset}
\begin{algorithmic}[1]
    \State \textbf{Input:} \textit{pred} \Comment{Raw prediction from the model}
    \State \textbf{Output:} \textit{cleansed\_pred} \Comment{Cleansed numerical prediction}

    \State Remove commas from \textit{pred}
    \State Extract all numbers from \textit{pred} using regex
    \State Select the first or last number based on context
    \State \Return \textit{cleansed\_pred}
\end{algorithmic}
\end{algorithm}
In the context of the GSM8K dataset, the answer-cleansing process is crucial for ensuring the accuracy and usability of predictions from large language models. Initially, the raw prediction, referred to as \texttt{pred}, often contains numerical answers formatted with commas or mixed with textual content. To standardize these predictions, we first remove any commas to normalize the numbers. Subsequently, we use regular expressions to extract all numerical values from this cleaned string. Given the nature of GSM8K tasks, where a specific numerical answer is typically required, our algorithm strategically selects either the first or last number based on predefined logic tailored to the dataset’s requirements. This selection process is designed to pick the most relevant number based on its position in the model's output. The final step produces a \texttt{cleansed\_pred}, which is the processed and formatted numerical answer ready for evaluation against the dataset's ground truth.
\end{document}